\pdfoutput=1

\documentclass[11pt]{article}

\usepackage[]{acl}

\usepackage{times}
\usepackage{latexsym}

\usepackage[T1]{fontenc}

\usepackage[utf8]{inputenc}

\usepackage{microtype}

\usepackage{inconsolata}

\usepackage{graphicx}

\usepackage{amssymb}
\usepackage{amsmath}
\usepackage{booktabs}

\usepackage{colortbl}  
\usepackage{xcolor}
\usepackage{array}

\definecolor{gg}{HTML}{e2f0cb}

\usepackage{multirow}

\usepackage{pifont}
\newcommand{\ccmark}{\ding{51}}%
\newcommand{\xxmark}{\ding{55}}%

\newcommand*{\affmark}[1][*]{\textsuperscript{#1}}
\usepackage{fdsymbol}

\usepackage[many]{tcolorbox}	
     
\tcbset{
    sharp corners,
    colback = white,
}   
\newtcolorbox{boxA}{
    rounded corners,
    arc = 5pt 
}

\title{Mixture-of-Subspaces in Low-Rank Adaptation}

\author{
Taiqiang Wu\affmark[$\diamondsuit$] \ Jiahao Wang\affmark[$\diamondsuit$] \ Zhe Zhao\affmark[$\spadesuit$] \ Ngai Wong\affmark[$\diamondsuit$]
\\
\affmark[$\diamondsuit$]The University of Hong Kong \ 
\affmark[$\spadesuit$]Tencent AI Lab
\\
{\tt \{takiwu,jiahao.wang\}@connect.hku.hk} \ {\tt nwong@eee.hku.hk}
}

\begin{document}
\maketitle

\begin{abstract}
In this paper, we introduce a \textit{subspace}-inspired Low-Rank Adaptation (LoRA) method, which is computationally efficient, easy to implement, and readily applicable to large language, multimodal, and diffusion models. 
Initially, we equivalently decompose the weights of LoRA into two subspaces, and find that simply mixing them can enhance performance. 
To study such a phenomenon, we revisit it through a fine-grained subspace lens, showing that such modification is equivalent to employing a fixed \textit{mixer} to fuse the subspaces. 
To be more flexible, we jointly learn the mixer with the original LoRA weights, and term the method as \textit{Mixture-of-Subspaces LoRA (MoSLoRA)}. 
MoSLoRA consistently outperforms LoRA on tasks in different modalities, including commonsense reasoning, visual instruction tuning, and subject-driven text-to-image generation, demonstrating its effectiveness and robustness.
Codes are available at \href{https://github.com/wutaiqiang/MoSLoRA}{github}.
\end{abstract}

\section{Introduction}

Large Language Models~(LLMs), such as GPT-4 \citep{DBLP:journals/corr/abs-2303-08774}, LLaMA 3 \citep{llama3modelcard}, and InternLM2 \citep{DBLP:journals/corr/abs-2403-17297}, have demonstrated remarkable performance across diverse disciplines \citep{DBLP:journals/corr/abs-2308-12950, thirunavukarasu2023large}.
Such strong capability is often attributed to the increased scale of training data and model parameters.
However, it also brings increasing challenges to adapting these LLMs for downstream tasks via fully fine-tuning all the parameters.

\begin{figure}[!t]
	\centering
 \includegraphics[width=0.95\linewidth]{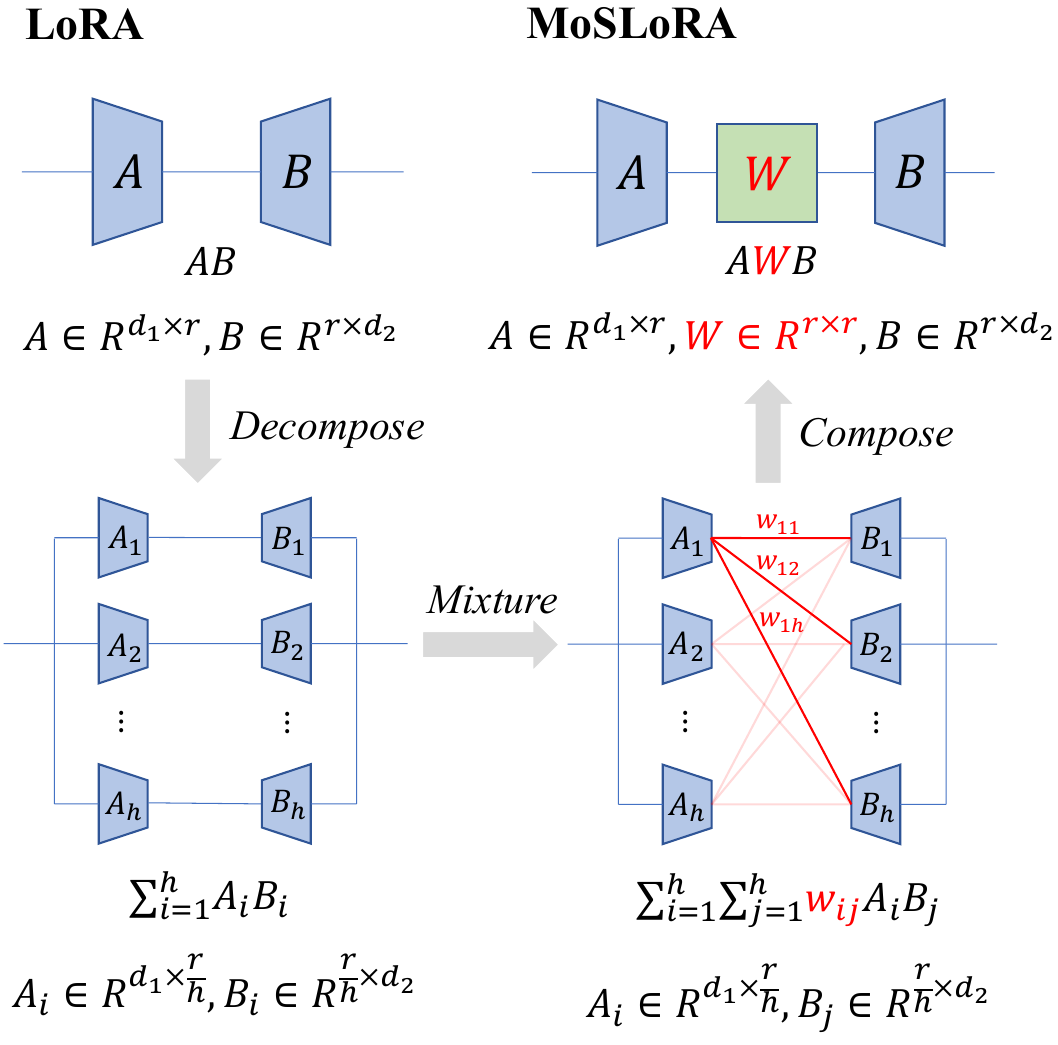}
	\caption{
	Comparison between vanilla LoRA and proposed MoSLoRA.
    In MoSLoRA, we employ learnable weights to mix more subspaces with negligible parameters~(i.e., $(d_1+d_2+r)r$ vs $(d_1+d_2)r$ and $d_1+d_2 \gg r$ typically).
    }
	\label{intro_fig}
\end{figure}

To tackle this issue, parameter-efficient fine-tuning~(PEFT) has been developed \citep{DBLP:conf/iclr/HuSWALWWC22, DBLP:conf/emnlp/LesterAC21, DBLP:conf/iclr/HeZMBN22} to minimize the number of optimized parameters while achieving comparable performance as much as possible.
Among these methods, LoRA~\citep{DBLP:conf/iclr/HuSWALWWC22} has gained increasing popularity due to its simplicity and efficacy, which proposes to update the extra low-rank branch exclusively and merge it into the frozen original weight during inference.
As shown in Figure \ref{intro_fig}, for the original weight matrix $\mathbf{W}_0 \in \mathbb{R}^{d_1 \times d_2}$,  the additional low-rank branch consists of a down projection $\mathbf{A} \in \mathbb{R}^{d_1 \times r}$ and an up projection $\mathbf{B} \in \mathbb{R}^{r \times d_2}$, where $r \ll \text{min}(d_1, d_2)$.
Hence, the number of updated parameters is reduced from $d_1 \times d_2$ to $(d_1+d_2)r$.

In this paper, we first define \emph{subspaces} in LoRA as the parallel components with smaller rank values, similar to the subspace in multi-head attention~(MHA) design \citep{DBLP:conf/nips/VaswaniSPUJGKP17}.
After that, we can decompose the vanilla LoRA into several subspaces via structural re-parameterization \citep{DBLP:journals/corr/abs-2305-09098, DBLP:conf/iccv/DingHT0HGD21}.
Figure \ref{mix-half-rank} indicates the process of decomposing into two subspaces.
Interestingly, we find that simply mixing these two subspaces performs better in the commonsense reasoning tasks.

Motivated by the observation, we further revisit the two-subspaces-mixing strategy in a more fine-grained~(rank=1) view and composed view.
In short, such a strategy equals inserting a \emph{mixer} matrix between $\mathbf{A}$ and $\mathbf{B}$, which is a fixed butterfly factor \citep{DBLP:conf/icml/DaoGERR19}.
Meanwhile, vanilla LoRA can be considered as a special case with a fixed identity matrix being the mixer.
Therefore, we propose MoSLoRA, a simple yet effective method, which employs a learnable mixer to fuse more subspaces and more flexibly.
As shown in Figure \ref{intro_fig}, we adapt the mixer $\mathbf{W}$ to fuse all the possible subspaces~(i.e., $A_iB_j$).
Compared to LoRA, MoSLoRA requires negligible extra parameters since $d_1+d_2 \gg r$.
Similarly to LoRA, MoSLoRA can also be merged into the original weights, and thus introduce no latency during inference.

We perform experiments on various downstream tasks, including commonsense reasoning tasks fine-tuning LLaMA 3 \citep{llama3modelcard}, visual instruction tuning on LLaVA-1.5 \citep{DBLP:journals/corr/abs-2310-03744} series models, and subject-driven text-to-image generation on Stable Diffusion XL~(SDXL) model \citep{DBLP:journals/corr/abs-2307-01952}.
Experimental results indicate that the proposed MoSLoRA consistently outperforms LoRA and other baselines, demonstrating its effectiveness and robustness.
Our contributions can be concluded as follows:
\begin{itemize}
    \item We decompose LoRA into subspaces via structural re-parameterization, revealing a new pathway to investigate LoRA.
    \item We propose a simple yet effective MoSLoRA method, employing a learnable mixer to fuse more subspaces and more flexibly. 
    \item We conduct extensive experiments on various downstream tasks, demonstrating the effectiveness and robustness of the proposed MoSLoRA.
\end{itemize}

\section{Preliminaries and Motivation}

\subsection{LoRA and Subspace View}

Based on the hypothesis that the update in weights during model adaptation exhibits low intrinsic rank, LoRA~\citep{DBLP:conf/iclr/HuSWALWWC22} aims to model the weight update via two low-rank matrices.
For a pre-trained weight matrix $\mathbf{W}_0 \in \mathbb{R}^{d_1 \times d_2}$ and arbitrary input $x$, they modify the forward pass as follows \footnote{In this paper, we use the post-multiplication for simplicity.}:
\begin{equation}
    x\mathbf{W}_0 + x\Delta \mathbf{W} = x\mathbf{W}_0+x\mathbf{A}\mathbf{B},
\end{equation}
where $\mathbf{A} \in \mathbb{R}^{d_1 \times r}$, $\mathbf{B} \in \mathbb{R}^{r \times d_2}$ and $r \ll \text{min}(d_1, d_2)$.
Typically, $\mathbf{A}$ is initialized as a Gaussian matrix and $\mathbf{B}$ as a zero matrix, so that $\Delta \mathbf{W}$ is zero at the beginning.
During training, the original weight $\mathbf{W}_0$ is frozen, while $\mathbf{A}$ and $\mathbf{B}$ contain trainable parameters.
After that, the $\mathbf{A}$ and $\mathbf{B}$ can be merged into $\mathbf{W}_0$ during inference, thus not introducing any latency.

\begin{figure}[!t]
	\centering
 \includegraphics[width=0.95\linewidth]{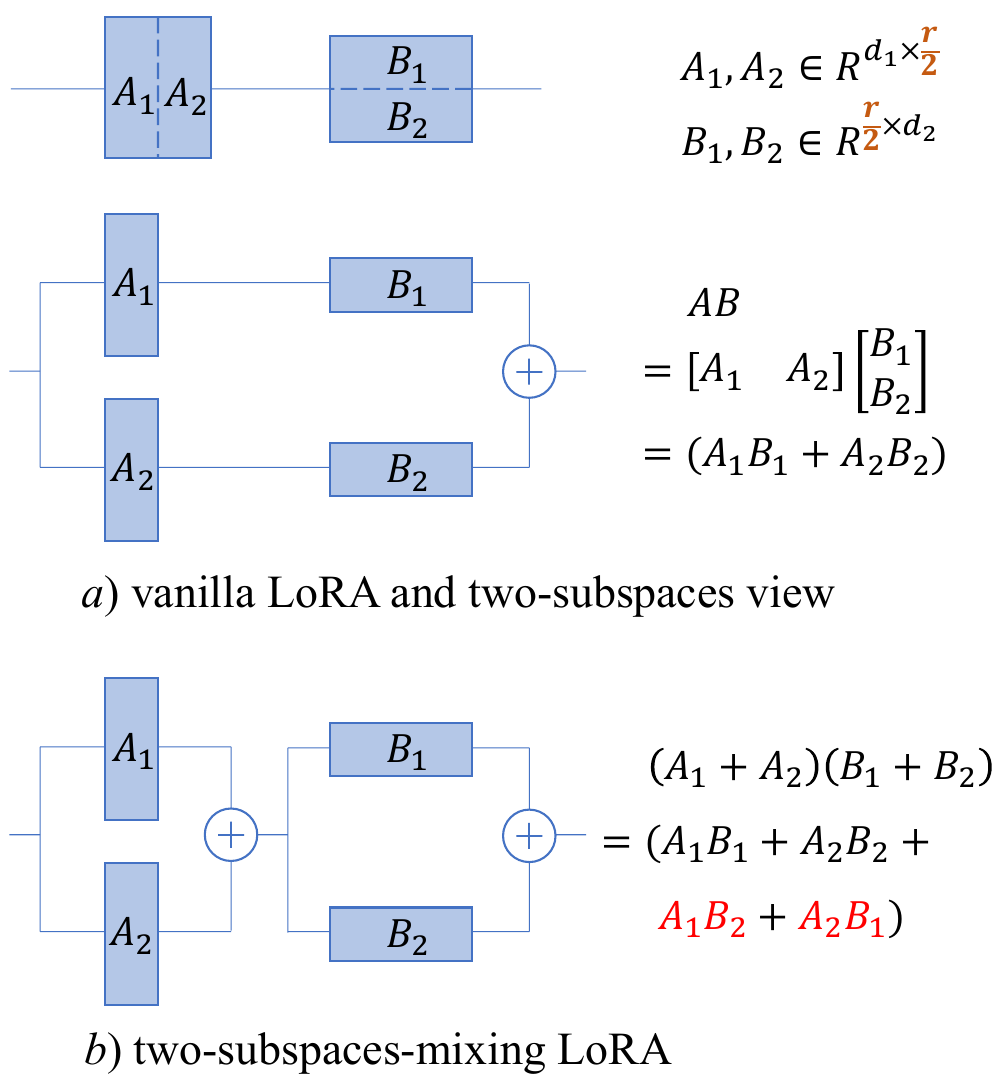}
	\caption{
	Overview of decomposing vanilla LoRA into two subspaces and mixing them. 
    Compared to vanilla LoRA, two-subspaces-mixing LoRA contains two extra entries.
    }
	\label{mix-half-rank}
\end{figure}

In this paper, we decompose LoRA into subspaces via structural re-parameterization, where the subspaces are defined as parallel components with smaller rank values. 
Figure \ref{mix-half-rank} part a shows the procedure for two subspaces.
Specifically, we decompose the $\mathbf{A}$ into two parts~(i.e., $\mathbf{A}_1$ and $\mathbf{A}_2$) by column, and $\mathbf{B}$ by row to get $\mathbf{B}_1$ and $\mathbf{B}_2$.
Therefore, we can easily get that:
\begin{equation}
\begin{aligned}
    x\mathbf{A}\mathbf{B} 
    & = x\begin{bmatrix} \mathbf{A}_1 & \mathbf{A}_2\end{bmatrix}
    \begin{bmatrix} \mathbf{B}_1 \\ \mathbf{B}_2\end{bmatrix} \\
    &= x(\mathbf{A}_1\mathbf{B}_1+\mathbf{A}_2\mathbf{B}_2),
\end{aligned}
\label{eq_2_2}
\end{equation}
where the $\mathbf{A}_1\mathbf{B}_1$ and $\mathbf{A}_2\mathbf{B}_2$ are the two subspaces.
In the two-subspace view, vanilla LoRA equals the sum of two subspaces.
Moreover, we can get a more fine-grained view if we split $\mathbf{A}$ and $\mathbf{B}$ for more parts, respectively.

\subsection{Mixing Two Subspaces}
\label{sec_mix_subspaces}

\begin{table*}[!t]
\centering
\resizebox{0.9\linewidth}{!}
{%
    \begin{tabular}{l|cccccccc|c}
        \toprule
        {\bf Method} & {\bf ARC-e} & {\bf OBQA} & {\bf SIQA} & {\bf ARC-c} & {\bf WinoG.} & {\bf PIQA} & {\bf BoolQ} & {\bf HellaS.} & {\bf Avg.} \\
        \midrule
        LoRA~(r=16)   &  87.7 & 82.8 & 79.3 & 75.7 & 84.8 & 86.7 & 72.3 & 93.5 & 82.8 \\
        + TS-Mixing & 88.3 & 83.0 & 80.3 & 78.1 & 84.8 & 87.5 & 73.8 & 94.3 & \textbf{83.8} \\
        \midrule
        LoRA~(r=32)   &  83.5 & 82.6 & 80.3 & 70.3 & 82.6 & 85.7 & 71.3 & 91.4 & 81.0 \\
        + TS-Mixing & 87.9 & 84.2 & 79.9 & 75.1 & 84.8 & 86.9 & 72.1 & 93.3 & \textbf{83.0} \\
        \bottomrule
    \end{tabular}
}
\caption{
Comparison of vanilla LoRA and two-subspaces-mixing LoRA~(denoted as TS-Mixing) on 8 benchmarks.
Simply mixing these two subspaces leads to better performance. 
}
\label{tab: 2_2_res}
\end{table*}

As shown in Figure \ref{mix-half-rank}b, we can simply mix two subspaces by adding up the outputs of $\mathbf{A}_1$ and $\mathbf{A}_2$.
Hence, the output of the whole module for input $x$ would be:
\begin{equation}
\begin{aligned}
&x(\mathbf{A}_1+\mathbf{A}_2)(\mathbf{B}_1+\mathbf{B}_2)
\\=&x
(
\mathbf{A}_1\mathbf{B}_1
+\mathbf{A}_2\mathbf{B}_2
+\textcolor{red}{\mathbf{A}_1\mathbf{B}_2}
+\textcolor{red}{\mathbf{A}_2\mathbf{B}_1}
).
\end{aligned}
\label{eq_2_4}
\end{equation}
Compared to Equation \ref{eq_2_2}, Equation \ref{eq_2_4} contains two extra entries and can model more information intuitively.

To compare these two strategies, we conduct experiments on the commonsense reasoning tasks following \citet{DBLP:conf/emnlp/HuWLXLB0PL23}.
We first fine-tune LLaMA-3 8B model \citep{llama3modelcard} on 170k training samples \citep{DBLP:conf/emnlp/HuWLXLB0PL23}, and then report the performance on 8 benchmarks, including ARC-c/e \citep{DBLP:journals/corr/abs-1803-05457}, OBQA \citep{DBLP:conf/emnlp/MihaylovCKS18}, SIQA \citep{DBLP:journals/corr/abs-1904-09728}, WinoG.~(WinoGrande) \citep{DBLP:conf/aaai/SakaguchiBBC20}, PIQA \citep{DBLP:conf/aaai/BiskZLGC20}, BoolQ \citep{DBLP:conf/naacl/ClarkLCK0T19}, and HellaS.~(HellaSwag) \citep{DBLP:conf/acl/ZellersHBFC19}.  
Please refer to Appendix \ref{appdix_data_detail_reason} for details of these benchmarks. 
All hyperparameters are the same and listed in Appendix \ref{appdix_hyper_detail_reason}.

Table \ref{tab: 2_2_res} shows the results on 8 benchmarks for these two methods.
Mixing two subspaces would lead to better performance under different settings~($r$=8/16), such as 93.3 compared to 91.4 of LoRA on the HellaSwag benchmark, showing the effectiveness and robustness of two-subspaces-mixing LoRA than vanilla LoRA.

\section{Methodology}

\subsection{More Fine-grained Subspace}

\begin{figure*}[!t]
	\centering
 \includegraphics[width=\linewidth]{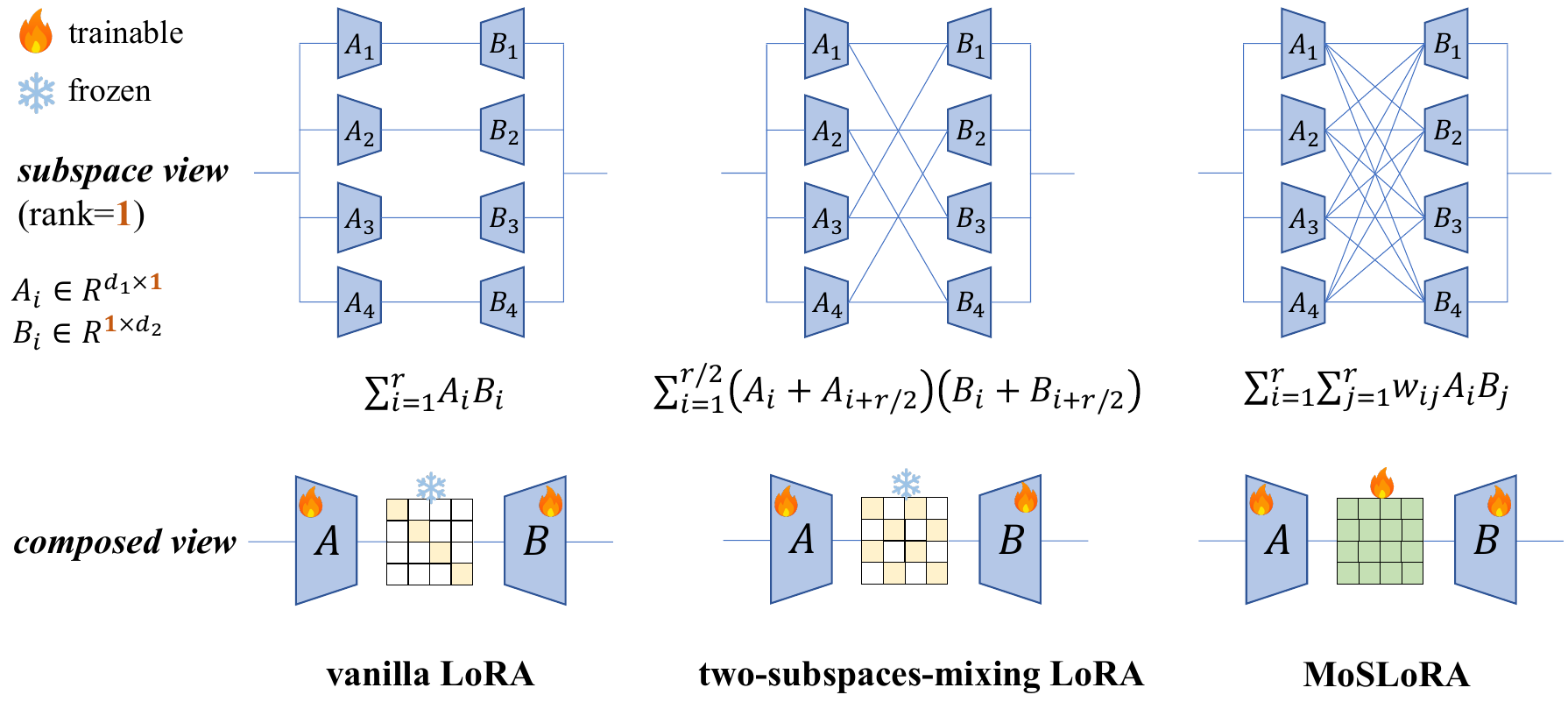}
	\caption{
	The subspace view~(rank=1) and composed view for vanilla LoRA, two-subspaces-mixing LoRA, and proposed MoSLoRA.
    In MoSLoRA, we employ a learnable mixer to fuse more information and more flexibly.
    }
	\label{mixer-view}
\end{figure*}

Motivated by the observation that mixing two subspaces would lead to better performance, we revisit the two-subspaces-mixing LoRA in view of more fine-grained subspace~(i.e., rank=1).
Specifically, we decompose the $\mathbf{A} \in \mathbb{R}^{d_1 \times r}$ and $\mathbf{B} \in \mathbb{R}^{r \times d_2}$ into $r$ subspaces~(rank=1), which can be formulated as:
\begin{equation}
\begin{aligned}
    \mathbf{A} &= \begin{bmatrix} \mathbf{A}_1 & \mathbf{A}_2 & \cdots & \mathbf{A}_r \end{bmatrix} \\
    \mathbf{B}^T &= \begin{bmatrix} \mathbf{B}_1^T & \mathbf{B}_2^T & \cdots & \mathbf{B}_r^T \end{bmatrix},
\end{aligned}
\end{equation}
where $\mathbf{A}_i \in \mathbb{R}^{d_1 \times 1}$ and $\mathbf{B}_i \in \mathbb{R}^{1 \times d_2}$ for $1\leq i \leq r$.
As shown in Figure \ref{mixer-view}, we can thus view vanilla LoRA as:
\begin{equation}
\begin{aligned}
    x\mathbf{A}\mathbf{B} = x\sum\limits_{i=1}^{r} \mathbf{A}_i\mathbf{B}_i = x\mathbf{A} \mathbf{I}_{r\times r} \mathbf{B}.
\end{aligned}
\label{eq_mix_vanilla}
\end{equation}
The $\mathbf{I}_{r\times r} \in \mathbb{R}^{r \times r}$ denotes the identity matrix.
Meanwhile, the two-subspaces-mixing LoRA equals to:
\begin{equation}
\begin{aligned}
    &x\sum\limits_{i=1}^{r/2} (\mathbf{A}_i+\mathbf{A}_{i+r/2})(\mathbf{B}_i+\mathbf{B}_{i+r/2}) \\
    = &x\mathbf{A} \begin{bmatrix} \mathbf{I}_{r/2\times r/2} & \mathbf{I}_{r/2\times r/2} \\
    \mathbf{I}_{r/2\times r/2} & \mathbf{I}_{r/2\times r/2} \end{bmatrix}
     \mathbf{B}.
\end{aligned}
\label{eq_mix_half}
\end{equation}

\begin{table}[!t]
\centering
\resizebox{\linewidth}{!}
{%
    \begin{tabular}{l|cc}
        \toprule
        {\bf Method} & {\bf \#N of subspaces~(rank=1)} &
        {\bf Trainable} \\
        \midrule
        LoRA & $r$ & \xxmark \\
        TS-Mixing & $2r$ & \xxmark \\
        \rowcolor{gg} MoSLoRA & $r^2$ & \ccmark \\
        \bottomrule
    \end{tabular}
}
\caption{
Comparison of LoRA, two-subspaces-mixing LoRA (denoted as TS-Mixing), and proposed MoSLoRA.
\#N denotes the number of mixed subspaces.
}
\label{tab: method-compare}
\end{table}

Interestingly, we can find that Equation \ref{eq_mix_vanilla} and Equation \ref{eq_mix_half} share the same paradigm:
\begin{equation}
\mathbf{A}\textcolor{red}{\mathbf{W}}\mathbf{B} ,
\end{equation}
where $\mathbf{W} \in \mathbb{R}^{r\times r}$ and we define $\mathbf{W}$ as the weight of \textbf{mixer} to fuse the subspaces.
For vanilla LoRA, the mixer is the fixed identity matrix fusing $r$ subspaces.
For the two-subspaces-mixing LoRA, the mixer is a fixed butterfly factor fusing $2r$ subspaces, which is more than LoRA.
Therefore, we propose MoSLoRA, adapting a trainable mixer to fuse all the possible subspaces.
As shown in Table \ref{tab: method-compare}, MoSLoRA mixes the information of $r^2$ subspaces~(rank=1) employing trainable weights, modeling the information of more subspaces and more flexible than LoRA.

\subsection{Initialization Strategies for Mixer}
\label{sec_mixer_init}
\begin{table}[!t]
\centering
\resizebox{\linewidth}{!}
{%
    \begin{tabular}{lc}
        \toprule
        {\bf Initialization Strategy} & {\bf Average Score}
        \\
        \midrule
        Zero Matrix & \emph{not converge} \\
        Identity Matrix & 82.6 \\
        Normal Distribution & 80.7 \\
        Orthogonal Matrix & 84.4 \\
        Kaiming Uniform Distribution & 85.6 \\ 
        \bottomrule
    \end{tabular}
}
\caption{
Comparison of various initialization strategies for the trainable mixer in MoSLoRA.
We report the average score on the commonsense reasoning tasks.
}
\label{tab: mixer_init_compare}
\end{table}

In the proposed MoSLoRA, we employ a trainable mixer to fuse all possible subspaces.
However, the system of MoSLoRA is linear, and a bad initialization hampers the learning \citep{DBLP:conf/iccv/HeZRS15}.
In MoSLoRA, we follow the setting in LoRA and initialize $\mathbf{A}$ using a \emph{Kaiming uniform} distribution \footnote{In the code of LoRA, they use Kaiming uniform initialization rather than Gaussian distribution claimed in the paper.} and $\mathbf{B}$ as a \emph{zero} matrix.
For the mixer weight $\mathbf{W}$, we compare various initialization strategies, including zero matrix, identity matrix, normal distribution, orthogonal matrix \citep{DBLP:journals/corr/SaxeMG13}, and Kaiming uniform distribution \citep{DBLP:conf/iccv/HeZRS15}.
Hyperparameters for finetuning can be found at Appendix \ref{appdix_hyper_detail_reason}.

Table \ref{tab: mixer_init_compare} reports the results of the commonsense reasoning tasks.
If we initialize the mixer as the zero matrix, then the model would not converge since all of the $\mathbf{A}$, $\mathbf{B}$, and $\mathbf{W}$ get zero gradients~(cf. Appendix \ref{appendix:mixer_zero} for proof).
When initializing the mixer as an identity matrix and updating it during training, the performance is similar to the vanilla LoRA with a fixed identity~(82.6 vs. 82.8).
Moreover, Kaiming uniform distribution and orthogonal matrix get strong performance, and thus we adapt them for the initialization of the mixer in MoSLoRA.

\subsection{Relation with Mixture-of-Experts}

Mixture-of-Experts~(MoE) methods aim to partition a set of parameters into experts and route input samples to specific experts during training and inference \citep{DBLP:journals/corr/abs-2209-01667, shi2024unchosen}.
Typically, they employ a router to generate scores for each expert based on the input, and then select top-k experts \citep{DBLP:journals/jmlr/FedusZS22, DBLP:conf/iclr/LepikhinLXCFHKS21, deepseekv2}.
In this paper, we propose MoSLoRA to mix the subspaces in LoRA, where the $w_{ij}$ in the mixer can be considered as the weight to compose subspace $\mathbf{A}_i\mathbf{B}_j$.
However, the differences between MoSLoRA and MoE methods are as follows:
\begin{itemize}
    \item In MoSLoRA, the weights to mix subspaces are input agnostic, while weights from gates in MoE methods are input specific. 
    \item In MoSLoRA, we adapt all the subspaces simultaneously, while MoE methods select top-k from all the experts.
\end{itemize}
\begin{table*}[!t]
\centering
\resizebox{\linewidth}{!}
{%
    \begin{tabular}{l|rrr|cccccccc|c}
        \toprule
        {\bf Method} & {\bf Param} & {\bf Time} & {\bf Mem} & {\bf ARC-e} & {\bf OBQA} & {\bf SIQA} & {\bf ARC-c} & {\bf WinoG.} & {\bf PIQA} & {\bf BoolQ} & {\bf HellaS.} & {\bf Avg.} \\
        \midrule
        LoRA & 28.3M & 8.0h & 29G &  87.7 & 82.8 & 79.3 & 75.7 & 84.8 & 86.7 & 72.3 & 93.5 & 82.8 \\
        LoKr & 0.9M & 26.3h & 66G & 89.2 & 81.8 & 78.7 & 76.7 & 82.1 & 81.6 & 65.1 & 92.0 & 80.9 \\
        LoHa & 28.3M & 25.5h & 68G & 91.2 & 85.8 & 81.1 & 80.5 & 83.3 & 89.7 & 75.0 & 95.0 & 85.2 \\
        FLoRA & 28.4M & 8.2h & 31G & 90.2 & 84.2 & 79.9 & 79.3 & 85.1 & 86.7 & 74.8 & 93.9 & 84.2 \\
        AdaLoRA & 28.3M & 12.5h & 58G & 90.4 & 85.0 & 76.7 & 79.1 & 83.3 & 86.4 & 75.1 & 75.4 & 81.4 \\
        DoRA & 29.1M & 14.5h & 33G & 90.1 & 87.2 & 80.3 & 79.1 & 84.7 & 88.8 & 74.5 & 95.5 & 85.0 \\
        DoRA$^*$ & 57.4M & 14.8h & 33G & 90.5 & 85.8 & 79.9 & 80.4 & 85.6 & 89.3 & 74.6 & 95.5 & 85.2 \\
        
        \rowcolor{gg} MoSLoRA & 28.4M & 8.2h & 31G & 90.5 & 86.8 & 81.0 & 81.5 & 85.8 & 89.7 & 74.6 & 95.0 & \textbf{85.6} \\
        \bottomrule
    \end{tabular}
}
\caption{
Accuracy comparison of various methods fine-tuning LLaMA-3 8B on the commonsense reasoning tasks.
\textbf{Param} denotes the number of trained parameters, \textbf{Time} for the training time on A100 GPU, and \textbf{Mem} for the GPU Memory usage.
$^*$ denotes a larger rank in DoRA.
We can find that the proposed MoSLoRA outperforms all the baselines with a slightly extra training cost than LoRA.
}
\label{tab: main_reason}
\end{table*}

\section{Experiments and Analysis}

\subsection{Commonsense Reasoning}

We fine-tune LLaMA-3 8B instruction version model \citep{llama3modelcard} for the commonsense reasoning question answering tasks.
We first train the model using 170k training samples \citep{DBLP:conf/emnlp/HuWLXLB0PL23}, and then test the fine-tuned model on 8 commonsense reasoning question answering benchmarks~(refer to Appendix \ref{appdix_data_detail_reason} for details).
The 170k training set is the mixture of the training sets of these benchmarks.
Besides LoRA~\citep{DBLP:conf/iclr/HuSWALWWC22}, we also compare MoSLoRA with various baselines, including:
1) LoKr \citep{DBLP:journals/corr/abs-2309-14859} which employs Kronecker products for matrix decomposition of $\mathbf{A}\mathbf{B}$;
2) LoHa \citep{DBLP:journals/corr/abs-2309-14859} which decomposes the vanilla LoRA into the Hadamard product of two LoRA branches;
3) FLoRA \citep{si2024flora} which introduces an extra core based on Tucker decomposition to maintain the consistent topological structure with the original space 
4) AdaLoRA \citep{DBLP:conf/iclr/ZhangCBH0CZ23} which parameterizes the incremental updates of the pre-trained
weight matrices in the form of singular value decomposition;
and 5) DoRA \citep{DBLP:journals/corr/abs-2402-09353} which decomposes the pretrained weight into its magnitude and directional components and fine-tunes both of them.

All the experiments are conducted using 1 Nvidia 80G A100 GPU.
The hyperparameters are listed in Appendix \ref{appdix_hyper_detail_reason}.
Based on the analysis in Table \ref{tab: mixer_init_compare}, we initialize the mixer following the Kaiming uniform distribution. 
Besides the accuracy, we also report the number of trained parameters and training overhead including time and peak GPU memory.

\begin{figure}[!t]
	\centering
\includegraphics[width=0.8\linewidth]{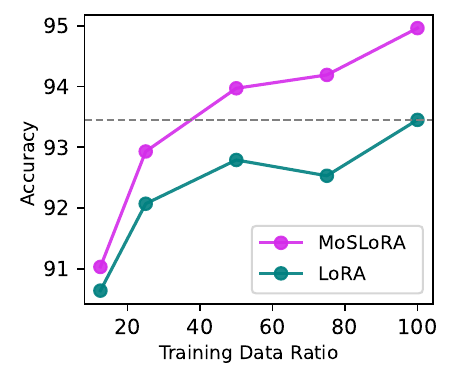}
\caption{
Comparison of MoSLoRA and LoRA on the HellaSwag benchmark with fewer training samples.
}
\label{fig: reason_few_samples}
\end{figure}

Table \ref{tab: main_reason} shows the results on 8 benchmarks.
Some findings can be summarized as follows:
\begin{itemize}
    \item MoSLoRA outperforms all the baselines, demonstrating the effectiveness of mixing the subspaces.
    Specifically, MoSLoRA gets an average of 85.6, which is 2.8 higher than the 82.8 of LoRA.
    Moreover, MoSLoRA outperforms DoRA with a higher rank.
    \item Compared to LoRA, MoSLoRA requires \emph{negligible} extra parameters~(less than 0.1M) and computing cost~(less than 0.2h).
    Meanwhile, MoSLoRA can save 44\% training time than DoRA and 68\% than LoHa.
    \item Though LoKr reduces the training parameters via Kronecher products, it requires more than 3x training time and 2x GPU memory than MoSLoRA.
    Also, LoKr gets an average score of 80.9, which is 4.7 lower than MoSLoRA.
\end{itemize}

\begin{table*}[!t]
\centering
\resizebox{\linewidth}{!}
{%
    \begin{tabular}{l|cc|ccccccccc|c}
        \toprule
        \multirow{2}{*}{{\bf Model}}
         & \multirow{2}{*}{{\bf Method}} & \multirow{2}{*}{{\bf Init.}} & \multicolumn{2}{c}{{\bf MMBench}} & {\bf SEED-} & \multirow{2}{*}{{\bf AI2D}} & {\bf SciQA} & {\bf Text} & {\bf Math} & {\bf MM-} & \multirow{2}{*}{{\bf MME}} & \multirow{2}{*}{{\bf Avg.}} \\
         &  &  & EN &  CN & {\bf Bench} &  & image & {\bf VQA} & {\bf Vista} & {\bf Vet} &  &  \\
         \midrule
        \multirow{3}{*}{LLaMA-3} & LoRA & - & 72.0 & 67.8 & 68.8 & 61.4 & 74.8 & 47.1 & 27.7 & 33.1 & 58.4 & 56.8 \\
        \cmidrule(lr){2-13}
        \multirow{2}{*}{+ViT} & \multirow{2}{*}{MoSLoRA} & \cellcolor{gg}{\emph{Orth}} & 73.0 & 68.2 & 69.0 & 61.2 & 75.7 & 47.2 & 27.6 & 33.4 & 60.6 & \cellcolor{gg}{\textbf{57.3}} \\
         &  & \cellcolor{gg}{\emph{Kai}} & 72.5 & 67.5 & 68.9 & 60.6 & 76.0 & 47.1 & 27.5 & 33.8 & 60.5 & \cellcolor{gg}{57.1} \\
        \midrule
         \multirow{3}{*}{InternLM2} & QLoRA & - & 70.8 & 68.9 & 70.4 & 62.2 & 72.5 & 49.8 & 30.2 & 33.9 & 61.6 & 57.8 \\
        \cmidrule(lr){2-13}
         \multirow{2}{*}{+ViT} & \multirow{2}{*}{QMoSLoRA} & \cellcolor{gg}{\emph{Orth}} & 73.5 & 71.2 & 71.1 & 64.8 & 71.8 & 49.8 & 30.2 & 35.0 & 62.0 & \cellcolor{gg}{58.8} \\
         &  & \cellcolor{gg}{\emph{Kai}} & 73.8 & 72.6 & 70.3 & 66.1 & 72.2 & 50.2 & 30.6 & 35.2 & 64.1 & \cellcolor{gg}{\textbf{59.5}} \\
        \bottomrule
    \end{tabular}
}
\caption{
Results on 9 benchmarks for vanilla LoRA and proposed MoSLoRA.
In MoSLoRA, we try both orthogonal~(denoted as \emph{Orth}) and Kaiming uniform initialization~(denoted as \emph{Kai}).
For InternLM2, we employ the 4-bit QLoRA on LoRA and MoSLoRA.
MoSLoRA consistently outperforms LoRA on various backbones for both initialization strategies.
}
\label{tab: main_visual}
\end{table*}

\paragraph{Fewer training samples} 
To compare the performance under fewer sample settings, we randomly select 12.5\%/25\%/50\%/75\% training samples from the original 170k training set and repeat the experiments.
As shown in Figure \ref{fig: reason_few_samples},
more training samples would lead to better performance and MoSLoRA outperforms LoRA under all the settings.
Particularly, MoSLoRA trained via 50\% samples gets a score of 83.6, which is 1.8 higher than LoRA using 100\% samples.
Moreover, the performance gap between MoSLoRA and LoRA becomes larger as the training samples increase, showing the superiority of MoSLoRA to modeling more complex information due to the mixture of subspaces.

\subsection{Visual Instruction Tuning}

To evaluate performance on multimodal tasks, we fine-tune the LLaVA-1.5 \citep{DBLP:journals/corr/abs-2310-03744} series models for visual instruction tuning, and then test the model for various visual QA benchmarks.

There are two stages in training LLaVA: 1) pretrain a two-layer MLP to project visual features to language space, and 2) optimize LLM and visual encoder~(optional) for visual instruction tuning.
In this paper, we employ the pretrained projector provided in XTuner \citep{2023xtuner}, and then conduct visual instruction tuning on the LLM backbone and visual encoder, simultaneously.
For the LLM backbones, we adapt the LLaMA3 8B \citep{llama3modelcard} and InternLM2 7B \citep{DBLP:journals/corr/abs-2403-17297} using the off-the-shelf projecters \footnote{\href{https://huggingface.co/xtuner}{pretrained projecters}}.
For the visual encoder, we employ the ViT \footnote{\href{https://huggingface.co/openai/clip-vit-large-patch14-336}{openai/clip-vit-large-patch14-336}}
\citep{DBLP:conf/iclr/DosovitskiyB0WZ21} large version.
Due to limited resources, we fintune both the LLM backbone and visual encoder via LoRA/MoSLoRA on the 665K instruction-following data \citep{DBLP:journals/corr/abs-2310-03744}, rather than optimize all the parameters in LLMs.
For InternLM2, we employ the 4-bit QLoRA \citep{DBLP:conf/nips/DettmersPHZ23} and corresponding QMoSLoRA~(QLoRA+MoSLoRA).
Based on the results in Table \ref{tab: mixer_init_compare}, we initialize the mixer as the orthogonal matrix and Kaiming uniform distribution, separately.
For specific hyperparameters, please refer to the Appendix \ref{appdix_hyper_detail_visual}.
It takes around 20 hours to fine-tune using 4 Nvidia A100 80G GPUs.

\begin{figure}[!t]
	\centering
\includegraphics[width=\linewidth]{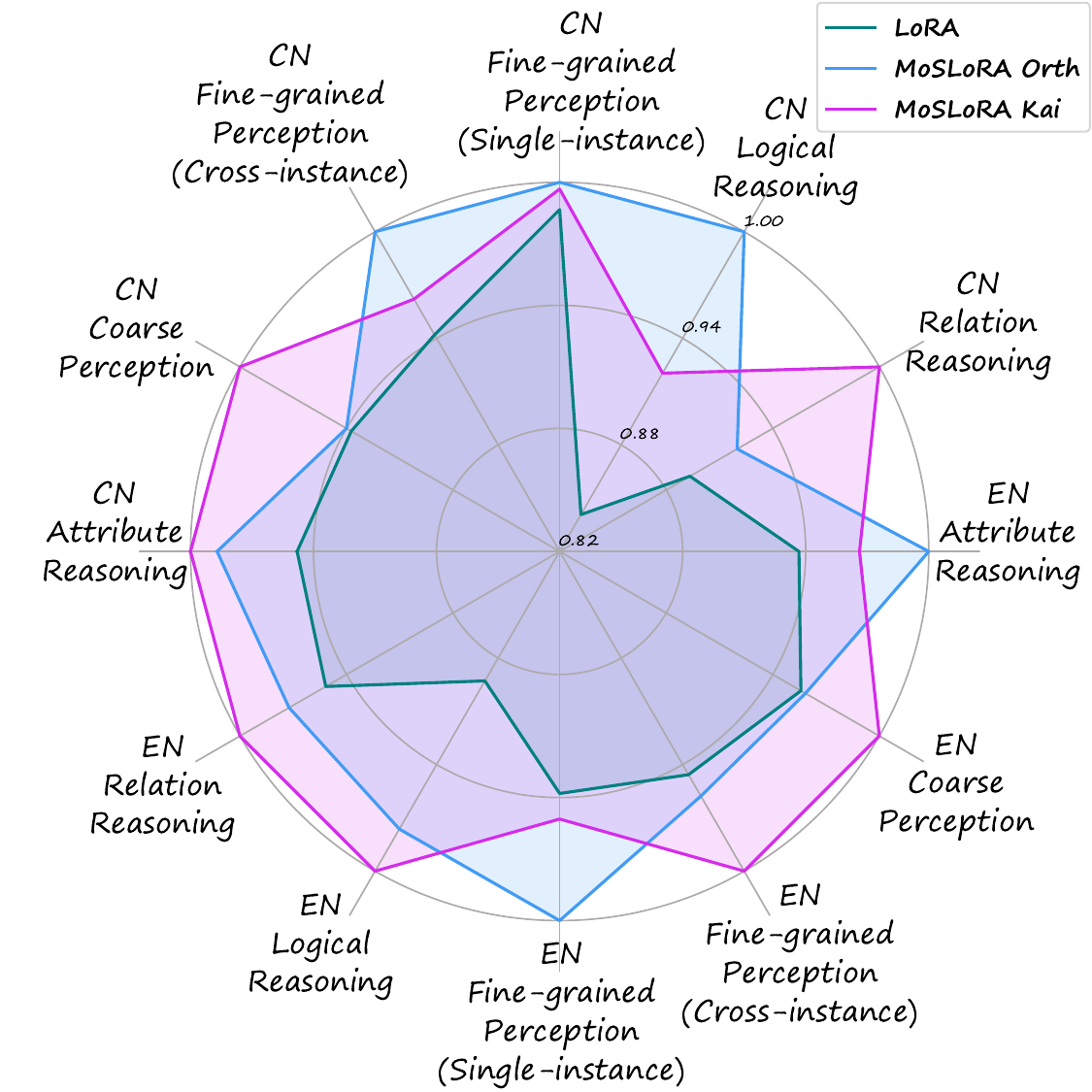}
\caption{
Normalized performance on 6 ability dimensions in MMBench EN/CN for QLoRA and QMoSLoRA when fintuning InternLM2.
MoSLoRA significantly improves the reasoning ability over LoRA.
}
\label{fig: mmbench_radar}
\end{figure}

\begin{figure*}[!t]
	\centering
\includegraphics[width=0.95\linewidth]{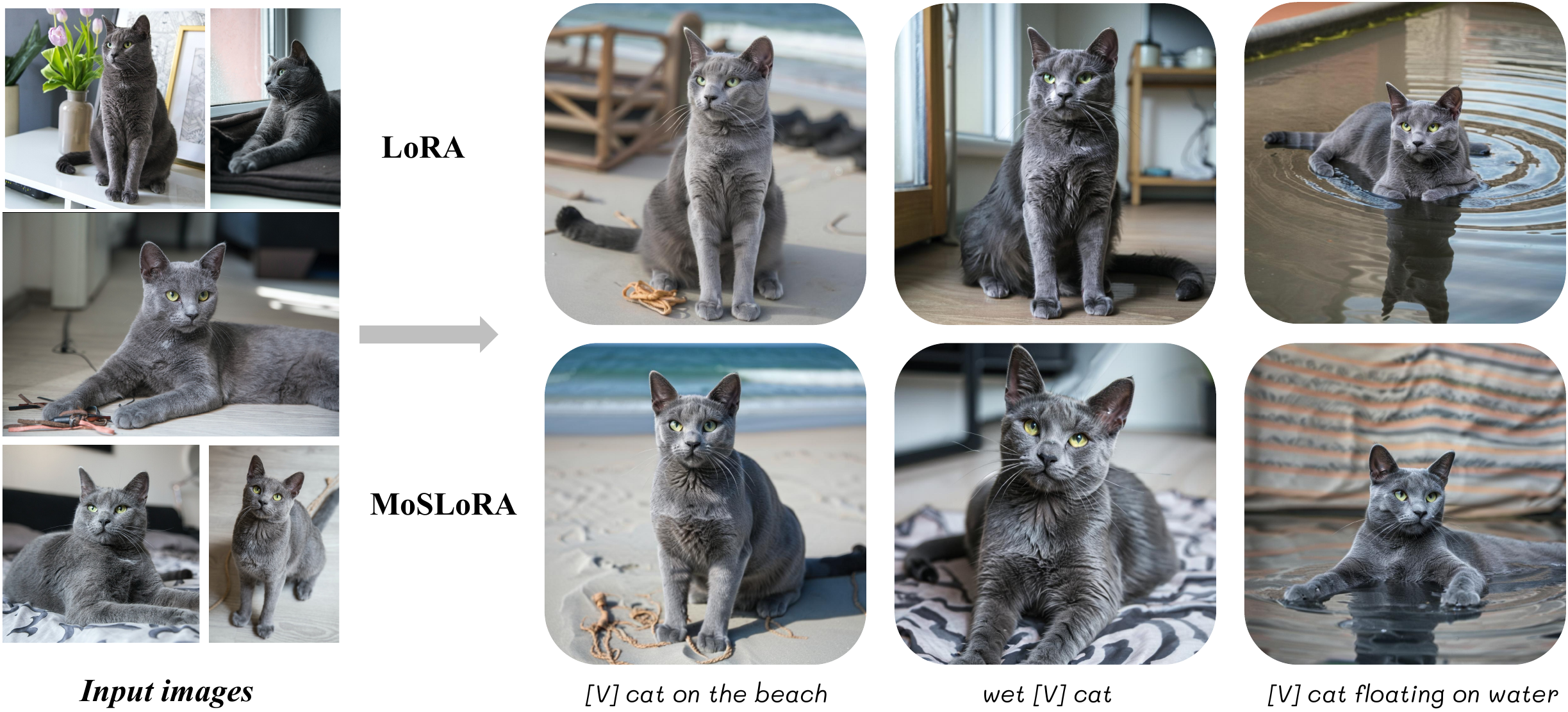}
\caption{
Comparison of generated images from LoRA and MoSLoRA on the subject-driven generation task.
MoSLoRA is more consistent with the subject in the input images~(e.g. the color of the hairs around the neck) and conforms to the given prompts~(e.g. the wet hair and floating gesture) better.
}
\label{fig: dreambooth_case}
\end{figure*}

After visual instruction tuning, we evaluate the trained model on 9 popular benchmarks, including MMBench EN/CN \citep{DBLP:journals/corr/abs-2307-06281}, SEED Bench \citep{DBLP:journals/corr/abs-2307-16125}, AI2D \citep{DBLP:conf/eccv/KembhaviSKSHF16}, SciQA \citep{lu2022learn}, TextVQA \citep{DBLP:conf/cvpr/SinghNSJCBPR19}, MathVista testmini \citep{DBLP:journals/corr/abs-2310-02255}, MM-Vet \citep{DBLP:journals/corr/abs-2308-02490}, and MME \citep{DBLP:journals/corr/abs-2306-13394}.
All the evaluations are done using the VLMEvalKit \citep{2023opencompass}.
Please refer to Appendix \ref{appdix_data_detail_visual} for the details of the dataset and the reported metrics.
Specifically, we scale the MME scores to 100 to calculate the average score.

Table \ref{tab: main_visual} shows the results on 9 benchmarks.
For both orthogonal and Kaiming initialization, MoSLoRA consistently outperforms LoRA on various benchmarks.
Specifically, MoSLoRA gets an average score of 59.5 on InternnLM2+ViT, which is 1.7 higher than LoRA.
Moreover, MoSLoRA also outperforms LoRA when combined with the 4-bit QLoRA.
It effectively showcases the compatibility of MoSLoRA with QLoRA.
Therefore, MoSLoRA can be applied in low-resource finetuning scenarios combined with the quantization methods.
In summary, the proposed MoSLoRA consistently outperforms LoRA in various settings, demonstrating its effectiveness and robustness.

\paragraph{More finegrained ability} 
Moreover, we also visualize the normalized scores on 6 ability dimensions in the MMbench EN/CN test set.
As shown in Figure \ref{fig: mmbench_radar}, we can observe that MoSLoRA performs better than LoRA on all abilities for both English and Chinese scenarios, especially the reasoning ability.
Reasoning tasks are typically considered to be more complex and difficult.
Compared to LoRA, MoSLoRA mixes more subspaces and is thus better at more difficult tasks such as logical reasoning.

\subsection{Subject-driven Generation}

We further perform the experiments fine-tuning the text-to-image diffusion models for the subject-driven generation task \citep{DBLP:conf/cvpr/RuizLJPRA23}.
The goal is to generate the images following the given prompts of one specific subject, which is defined in a few given images.
We first fine-tune a text-to-image model with the input images paired with a text prompt containing a unique identifier~(e.g., A photo of a [V] cat).
After that, we can employ other prompts containing the unique identifier to generate the corresponding images.

Figure \ref{fig: dreambooth_case} shows one case of a cat from the DreamBooth dataset \citep {DBLP:conf/cvpr/RuizLJPRA23}.
We finetune the SDXL\footnote{\href{https://huggingface.co/stabilityai/stable-diffusion-xl-base-1.0}{stable-diffusion-xl-base-1.0}} model \citep{DBLP:journals/corr/abs-2307-01952} via LoRA and MoSLoRA.
In MoSLoRA, the mixer is initialized as an orthogonal matrix.
During finetuning, the learning rate is 1e-4, and the batch size is 4.
We train the model for 500 steps, which costs around 16 minutes using 1 80G A100 GPU.
During generation, we infer 50 steps for the given prompts.
Compared to vanilla LoRA, we can find that our proposed MoSLoRA captures more details of the subject and better conforms to the given prompt.
Specifically, MoSLoRA learns more details about the given cat, including the color of the hairs around the neck and the shape of the paw.
Meanwhile, the images from MoSLoRA are more consistent with the given prompts, such as the wet~(thus clumped) hair and the floating gesture~(spread hands).

\begin{table}[!t]
\centering
\resizebox{\linewidth}{!}
{%
    \begin{tabular}{l|ccc|c}
        \toprule
        {\bf Metric} & {\bf Win} &
        {\bf Tie} & {\bf Loss} & $\Delta$ \\
        \midrule
        \emph{Sub-simi} & 23.1\% & 60.4\% & 16.5\% & +6.6\% \\
        \emph{Pro-cons} & 45.1\% & 44.2\% & 10.7\% & +34.3\%\\
        \bottomrule
    \end{tabular}
}
\caption{
Human evaluation results on the generated images comparing MoSLoRA against LoRA.
\emph{Sub-simi} denotes for the subject similarity and \emph{Pro-cons} for prompt consistency. 
}
\label{tab: sd-user}
\end{table}

\paragraph{Human evaluation} 

We also perform human evaluation on the generated images.
First, we choose four subjects~(i.e., cat, dog, grey sloth plushie, and can) from the DreamBooth dataset \citep {DBLP:conf/cvpr/RuizLJPRA23} and fine-tune the SDXL model, respectively.
Then, we randomly select 8 prompts to generate the corresponding images.
After that, 15 human experts are asked to independently score win/tie/loss for the paired images from LoRA and MoSLoRA.
During evaluation, we shuffle these pairs and keep that these experts do not know the source model of each image.
We employ two metrics, including 1)  \emph{subject similarity} defined as the similarity between subjects from generated images and given images, and 2) \emph{prompt consistency} defined as the consistency among prompts and generated images.
Table \ref{tab: sd-user} reports the average score for all the images.
We can find that MoSLoRA outperforms LoRA on both metrics.
In particular, MoSLoRA gets an average winning ratio of 45.1\% on prompt consistency, which is 34.3\% than LoRA.
Please refer to Appendix \ref{appendix:case_of_image} for the detailed prompts and corresponding generated images from LoRA and MoSLoRA.
\section{Related Work}

\subsection{Parameter-Efficient Fine-tuning} 
Parameter-efficient fine-tuning~(PEFT), aiming to update a small proportion of parameters to adapt Large Language Models~(LLMs), has become increasingly important.
The mainstreaming PEFT methods can be categorized into: 1) adapter based methods~\citep{DBLP:conf/icml/HoulsbyGJMLGAG19, DBLP:conf/nips/LeiBBALZ0ZWLZC23}, which inserts modules between transformer layers; 2) prefix tuning methods~\citep{DBLP:conf/acl/LiL20, DBLP:journals/corr/abs-2103-10385}, which prepends tunable prefix vectors into the hidden states; 3) selective methods~\citep{DBLP:conf/acl/ZakenGR22}, which select part of the parameters to update; and 4) low-rank adapting~(LoRA) series~\citep{DBLP:conf/iclr/HuSWALWWC22, DBLP:journals/corr/abs-2309-14859}, which injects trainable low-rank branches to approximate the weight updates.
In LoRA, low-rank branches can be merged into the original weights during inference, thus bringing no latency.
We refer the reader to \citet{DBLP:journals/corr/abs-2403-14608} for a more comprehensive survey.
In this paper, we focus on LoRA methods.

\subsection{LoRA and its Variants}
The core of LoRA is to update the mergeable and low-rank branches to model the weight updates.
\citet{DBLP:conf/iclr/HuSWALWWC22} initialize the branch as a product of two low-rank matrices.
The following variants can be categorized into: 1) \emph{introducing training skills}, such as setting different learning rates \citep{DBLP:journals/corr/abs-2402-12354} and adding random noise \citep{DBLP:journals/corr/abs-2404-09610}; 2) \emph{searching ranks}, such as DyLoRA \citep{DBLP:conf/eacl/ValipourRKG23} and AdaLoRA \citep{DBLP:conf/iclr/ZhangCBH0CZ23}; and 3) \emph{new designs} for the branch, such as LoKr \citep{DBLP:journals/corr/abs-2309-14859}, LoHa \citep{DBLP:journals/corr/abs-2309-14859}, VeRA \citep{DBLP:journals/corr/abs-2310-11454}, DoRA \citep{DBLP:journals/corr/abs-2402-09353}, and FLoRA \citep{si2024flora}.
LoKr and LoHa employ Kronecker and Hadamard products to replace the vanilla matrix product, respectively.
DoRA decomposes the pretrained weight into its magnitude and directional components and fine-tunes them separately.

We also notice a very recent concurrent work FLoRA \citep{si2024flora}.
Differences between MoSLoRA and FLoRA are as follows: 
1) \textit{initialization methods and the corresponding motivation}: 
The motivation of FLoRA is to maintain the structural integrity of the original high-dimensional parameter spaces~(i.e., 2D Convolution) by introducing the core spaces.
Differently, MoSLoRA is motivated by the observation of exploratory experiments in Section \ref{sec_mix_subspaces}, where we find the two-subspaces-mixing LoRA in the view of more fine-grained subspace has the potential to learn complex features. 
As a result, we introduce an extra learnable mixer initialized as Kaiming uniform distribution or orthogonal matrix, which we empirically~(Section \ref{sec_mixer_init}) find vital for the final performance. 
2) \textit{analysis and derivation}: 
The design of our mixer comes from the process of analyzing and deriving the two-subspaces-mixing LoRA.
We revisit the two-subspaces-mixing strategy in a more fine-grained~(rank=1) view and composed view, and then we find that both LoRA and two-subspaces-mixing LoRA are equivalent to inserting fixed mixer matrices.
Based on that, we propose MoSLoRA, which employs a learnable mixer to fuse more subspaces and more flexibly.
3) \textit{models and datasets}: 
For visual-instruction tuning tasks, FLoRA fine-tunes LLaVA-1.5-7B~(based on Vicuna-1.5-7B \citep{DBLP:journals/corr/abs-2304-03277}) and evaluates on seven vision language benchmarks: VQA$^{v2}$ \citep{DBLP:journals/ijcv/GoyalKASBP19}, GQA \citep{DBLP:conf/cvpr/HudsonM19}, VisWiz \citep{DBLP:conf/cvpr/Gurari0SGLGLB18}, SQA \citep{DBLP:conf/nips/LuMX0CZTCK22}, TextVQA \citep{DBLP:conf/cvpr/SinghNSJCBPR19}, POPE \citep{DBLP:conf/emnlp/LiDZWZW23}, and
MMBench \citep{DBLP:journals/corr/abs-2307-06281}.
Differently, We select LLaMA3 8B and InternLM2 as the language models and evaluate the fine-tuned models on MMBench EN/CN \citep{DBLP:journals/corr/abs-2307-06281}, SEED Bench \citep{DBLP:journals/corr/abs-2307-16125}, AI2D \citep{DBLP:conf/eccv/KembhaviSKSHF16}, SciQA \citep{lu2022learn}, TextVQA \citep{DBLP:conf/cvpr/SinghNSJCBPR19}, MathVista testmini \citep{DBLP:journals/corr/abs-2310-02255}, MM-Vet \citep{DBLP:journals/corr/abs-2308-02490}, and MME \citep{DBLP:journals/corr/abs-2306-13394}.
Besides, for language models, we evaluate MoSLoRA on LLaMA3 8B, while FLoRA focuses on DeBERTaV3 \citep{DBLP:conf/iclr/HeGC23}. 




 

\section{Conclusion}

This work proposes a novel MoSLoRA method for parameter-efficient fine-tuning.
We first decompose the LoRA into subspaces and find that simply mixing the half-rank subspaces would lead to better performance.
After that, we revisit vanilla LoRA and two-subspaces-mixing strategy in a more fine-grained view~(i.e., rank=1), thus unifying both methods as employing an extra fixed mixer.
Therefore, we propose MoSLoRA, which employs a learnable mixer to fuse more information and more flexibly.
The mixer requires negligible extra parameters and computing costs.
Experimental results on commonsense reasoning tasks, visual instruction tuning tasks, and subject-driven generation tasks demonstrate the effectiveness and robustness of the proposed MoSLoRA.
For future work, we would consider applying MoSLoRA for more tasks.
Finding a task-specific way to initialize the mixer for faster convergence would be another interesting topic.

\section*{Limitations}
In this paper, we conduct experiments on commonsense reasoning tasks, visual instruction tuning tasks, and subject-driven generation tasks.
LoRA can be applied in more scenarios, such as mixing styles in image generation tasks when fine-tuning stable diffusion models.
We leave these tasks for future work.

\section*{Ethics Statement}
This project aims to improve the LoRA methods and can be employed for subject-driven text-to-image generation tasks, where the users can fine-tune the stable diffusion models to generate images of a specific subject defined by the input images.
In some cases, such malicious parties might use the generated images to mislead viewers.
This is a common issue in generative model approaches or content manipulation techniques.

\section*{Acknowledgements}
We thank all anonymous reviewers for their constructive feedback on improving our paper.
We thank Zenan Xu, Chaofan Tao, Chenchen Ding, Shuqi Wang, and Zhengwu Liu for their fruitful discussions.
This work was supported by the Theme-based Research Scheme (TRS) project T45-701/22-R of the Research Grants Council (RGC), Hong Kong SAR.


\clearpage
\appendix

\section{Details of Benchmarks}

\subsection{Commonsene Reasoning}
\label{appdix_data_detail_reason}
The details of the benchmarks are as follows:
\begin{itemize}
\item ARC-c/e \citep{DBLP:journals/corr/abs-1803-05457}: the Challenge Set and Easy
Set of ARC dataset of genuine grade-school level, containing 2376/1172 multiple-choice science questions in the test set, respectively.
    
\item OBQA \citep{DBLP:conf/emnlp/MihaylovCKS18}: questions requiring multi-step reasoning, use of additional commonsense knowledge, and rich text comprehension. 
There are 500 questions in the test set.

\item SIQA \citep{DBLP:journals/corr/abs-1904-09728}: reasoning questions about people’s actions and their social implications.
There are 1954 questions in the test set.

\item WinoG.~(WinoGrande) \citep{DBLP:conf/aaai/SakaguchiBBC20}: fill-in-a-blank task with binary options to choose the right option for a given sentence which
requires commonsense reasoning.
There are 1267 questions in the test set.

\item PIQA \citep{DBLP:conf/aaai/BiskZLGC20}:
questions with two solutions requiring
physical commonsense.
There are 1830 questions in the test set.

\item BoolQ \citep{DBLP:conf/naacl/ClarkLCK0T19}:
yes/no questions which are naturally
occurring and generated in unprompted and unconstrained settings.
There are 3270 questions in the test set.

\item HellaS.~(HellaSwag) \citep{DBLP:conf/acl/ZellersHBFC19}: commonsense NLI questions including a context and several endings which complete the context.
There are 10042 questions in the test set.

\end{itemize}

For all the benchmarks, we report the accuracy following \citet{DBLP:conf/emnlp/HuWLXLB0PL23}.

\subsection{Visual Instruction Tuning}
\label{appdix_data_detail_visual}
The details of benchmarks and reported metrics are as follows:
\begin{itemize}
    \item MMBench EN/CN \citep{DBLP:journals/corr/abs-2307-06281}: the English and Chinese version of MMBench.
    MMBench contains over 3000 multiple-choice questions covering 20 different ability dimensions.
    Each ability dimension encompasses over 125 questions.
    We report the accuracy of the \emph{test} set 
\footnote{\href{https://mmbench.opencompass.org.cn/mmbench-submission}{Online submission for results}}.
    \item SEED Bench \citep{DBLP:journals/corr/abs-2307-16125}: 19K multiple choice questions with accurate human annotations, which spans 12 evaluation dimensions including the comprehension of both the image and video modality.
    In this paper, we use the image modality only and report the accuracy.
    \item AI2D \citep{DBLP:conf/eccv/KembhaviSKSHF16}: AI2 Diagrams~(AI2D) of over 5000 grade school science diagrams and more than 15000 corresponding multiple choice questions.
    We report the accuracy of the test set.
    \item SciQA~(ScienceQA) \citep{lu2022learn}: 21k multimodal multiple choice questions with diverse science topics and annotations of their answers with corresponding lectures and explanations.
    We report the accuracy of the test set.
    \item TextVQA \citep{DBLP:conf/cvpr/SinghNSJCBPR19}: 45,336 questions on 28,408 images that require reasoning about text to answer.
    We report the accuracy of the validation set.
    \item MathVista testmini \citep{DBLP:journals/corr/abs-2310-02255}: a benchmark designed to combine challenges from diverse mathematical and visual tasks. 
    It consists of 6,141 examples, derived from 28 existing multimodal datasets involving mathematics and 3 newly created datasets.
    We report the accuracy scores on the testmini subset of 1,000 examples using GPT-4-turbo.
    
    \item MM-Vet \citep{DBLP:journals/corr/abs-2308-02490}:
    200 images and 218 questions (samples), including 187 images from various online sources with 205 questions, 10 images from VCR with 10 paired questions, and 3 paired questions and images for medical expert knowledge.
    We report the average scores from the GPT-4-turbo.
    
    \item MME \citep{DBLP:journals/corr/abs-2306-13394}: 14 subtasks aiming to measure both perception and cognition abilities and the answer is yes or no.
    For the metrics, original scores include accuracy and accuracy+ for each task, and the total score is 2800.
    In this paper, we \emph{scale} the scores to 100 for average.
    
\end{itemize}

\begin{table*}[!t]
\centering
\resizebox{0.85\linewidth}{!}
{%
    \begin{tabular}{c|cccccccc}
        \toprule
        {\bf Hyperparameter} & LoRA & 
        LoKr &
        LoHa &
        FLoRA &
        AdaLoRA &
        MoSLoRA &
        DoRA &
        DoRA$^*$
        \\
       \midrule
        Rank r & \multicolumn{7}{c|}{16} & 32 \\
        \midrule
        $\alpha$ & \multicolumn{7}{c|}{32} & 64 \\
        \midrule
        Dropout & \multicolumn{8}{c}{0.05} \\
        \midrule
        Batch size & \multicolumn{8}{c}{16} \\
        \midrule
        Epochs & \multicolumn{8}{c}{3} \\
        \midrule
        Learning rate & \multicolumn{6}{c|}{3e-4} & \multicolumn{2}{c}{1e-4} \\
        \midrule
        Target module & \multicolumn{8}{c}{q, k, v, up, down} \\
        \bottomrule
    \end{tabular}
}
\caption{
The hyperparameters for various methods on the commonsense reasoning tasks.
}
\label{tab: hyper_commonsense}
\end{table*}
 
\begin{table}[!t]
\centering
\resizebox{0.95\linewidth}{!}
{%
    \begin{tabular}{c|cc}
        \toprule
        {\bf Hyperparameter} & LLaMA-3+ViT & InternLM2+ViT \\
       \midrule
        Batch size & 8 & 16 \\
        \midrule
        Accumulative & 2 & 1 \\
        \midrule
        Learning rate & \multicolumn{2}{c}{2e-5} \\
        \midrule
        Epoch & \multicolumn{2}{c}{1} \\
        \midrule
        Rank r & 64/64 & 512/64 \\
        \midrule
        $\alpha$ & 128/16 & 256/16 \\
        \midrule
        Target module & \multicolumn{2}{c}{q, k, v, o, up, down, gate} \\ 
        \bottomrule
    \end{tabular}
}
\caption{
The hyperparameters for various methods for visual instruction tuning.
For rank and alpha, we report in the format of LLM/Visual Encoder.
}
\label{tab: hyper_visual}
\end{table}

\section{Experimental Setup}

\subsection{Commonsene Reasoning}
\label{appdix_hyper_detail_reason}

Table \ref{tab: hyper_commonsense} shows the detailed hyper-parameters for commonsense reasoning tasking when fine-tuning the LLaMA3-8B instruction version.
For AdaLoRA, we set both the initial rank and target rank to be 16.

\subsection{Visual Instruction Tuning}
\label{appdix_hyper_detail_visual}

Table \ref{tab: hyper_visual} reports the detailed hyper-parameters for visual instruction tuning when fine-tuning the LLaMA3-8B+ViT and InternLM2+ViT.
Moreover, we employ the 4-bit QLoRA when finetuning the InternLM2, where the quantization type is NF4 with double quantization skills.

\section{Initialize Mixer as Zero Matrix}
\label{appendix:mixer_zero}

In MoSLoRA, we model the forward process as:
\begin{equation}
\begin{aligned}
    y&=x\mathbf{W}_{merge} \\
    \mathbf{W}_{merge} &= \mathbf{W}_0+\mathbf{A}\mathbf{W}\mathbf{B},  
\end{aligned}
\end{equation}
where the $\mathbf{W}_0$ is frozen during training.
Then we have:
\begin{equation}
\begin{aligned}
    \frac{\partial y}{\partial \mathbf{A}}&=
    \frac{\partial y}{\partial \mathbf{W}_{merge}} \mathbf{B}^T \mathbf{W}^T \\
    \frac{\partial y}{\partial \mathbf{W}}&=\mathbf{A}^T 
    \frac{\partial y}{\partial \mathbf{W}_{merge}} \mathbf{B}^T
    \\
    \frac{\partial y}{\partial \mathbf{B}}&=\mathbf{W}^T\mathbf{A}^T 
    \frac{\partial y}{\partial \mathbf{W}_{merge}} 
\end{aligned}
\label{eq: all_gradients}
\end{equation}

If we initialize $\mathbf{W}$ and $\mathbf{B}$ as zero matrices simultaneously, all the gradients in Equation \ref{eq: all_gradients} would be zero, and neither would be updated.



\section{Cases of Generated Images}
\label{appendix:case_of_image}

\begin{figure*}[!t]
	\centering
\includegraphics[width=\linewidth]{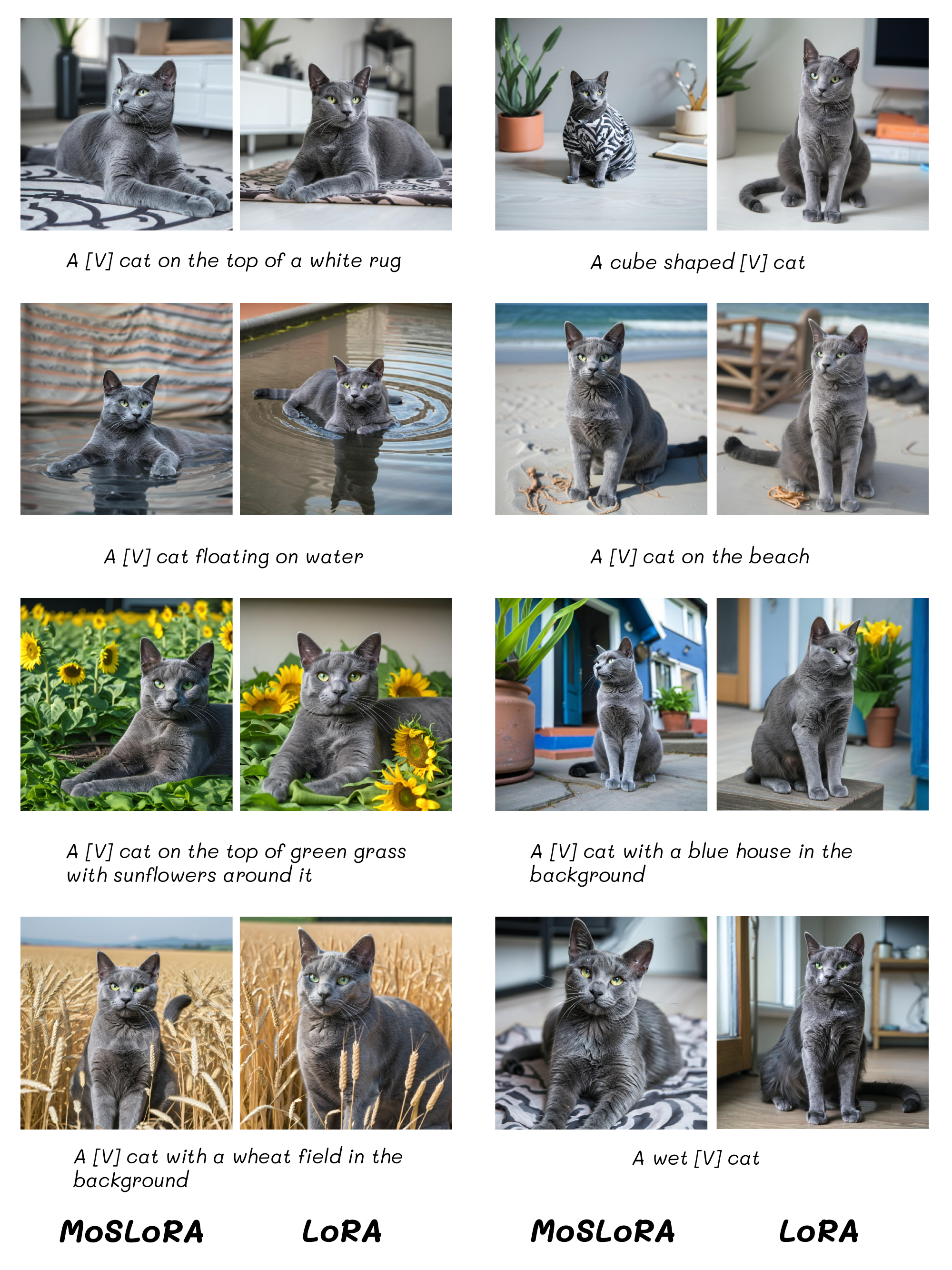}
\caption{
Cases of generated images and paired prompts for the subject \emph{cat}. 
}
\label{fig: appendix_sd_cases1}
\end{figure*}

\begin{figure*}[!t]
	\centering
\includegraphics[width=\linewidth]{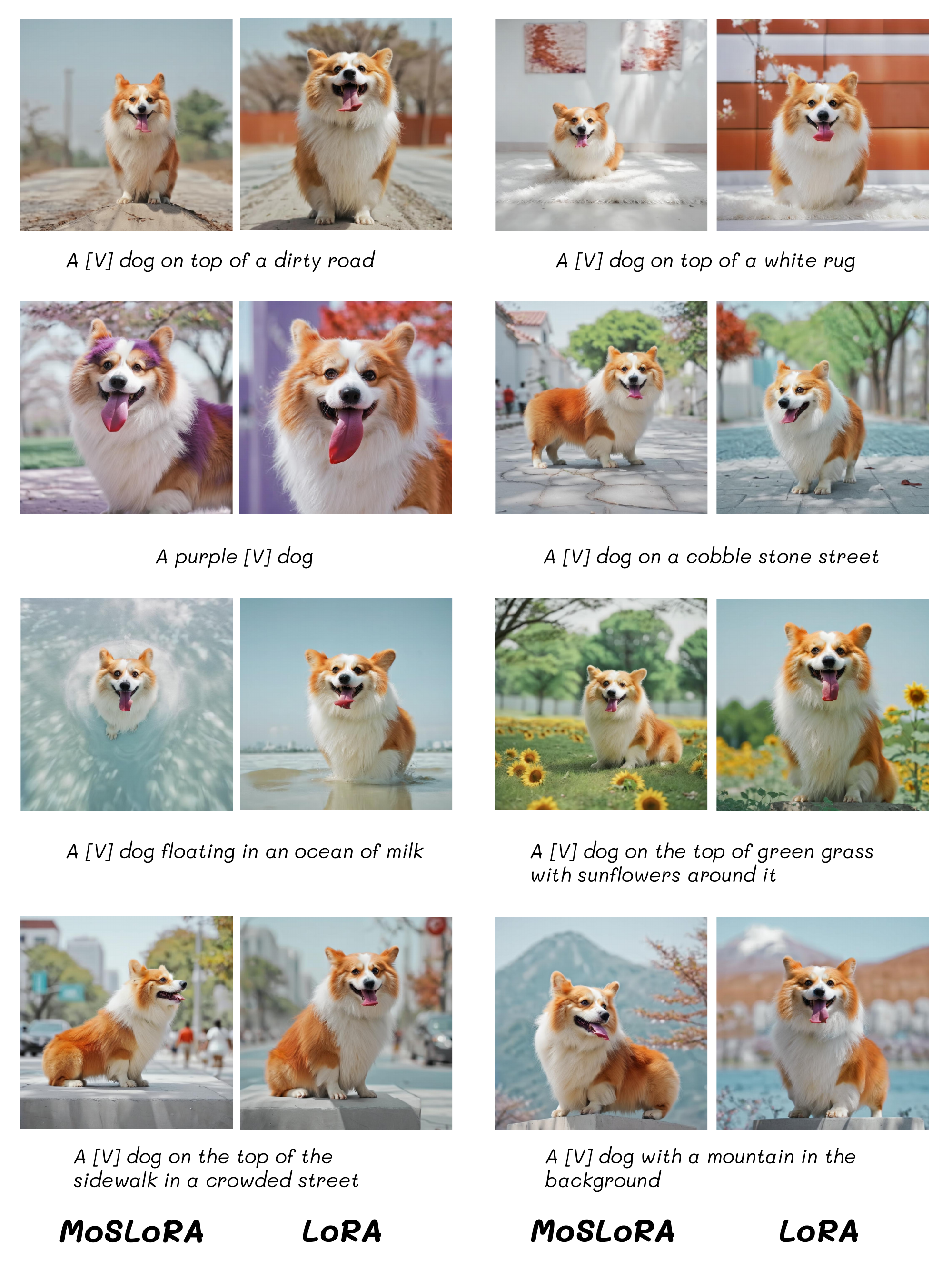}
\caption{
Cases of generated images and paired prompts for the subject \emph{dog}. 
}
\label{fig: appendix_sd_cases2}
\end{figure*}

\begin{figure*}[!t]
	\centering
\includegraphics[width=\linewidth]{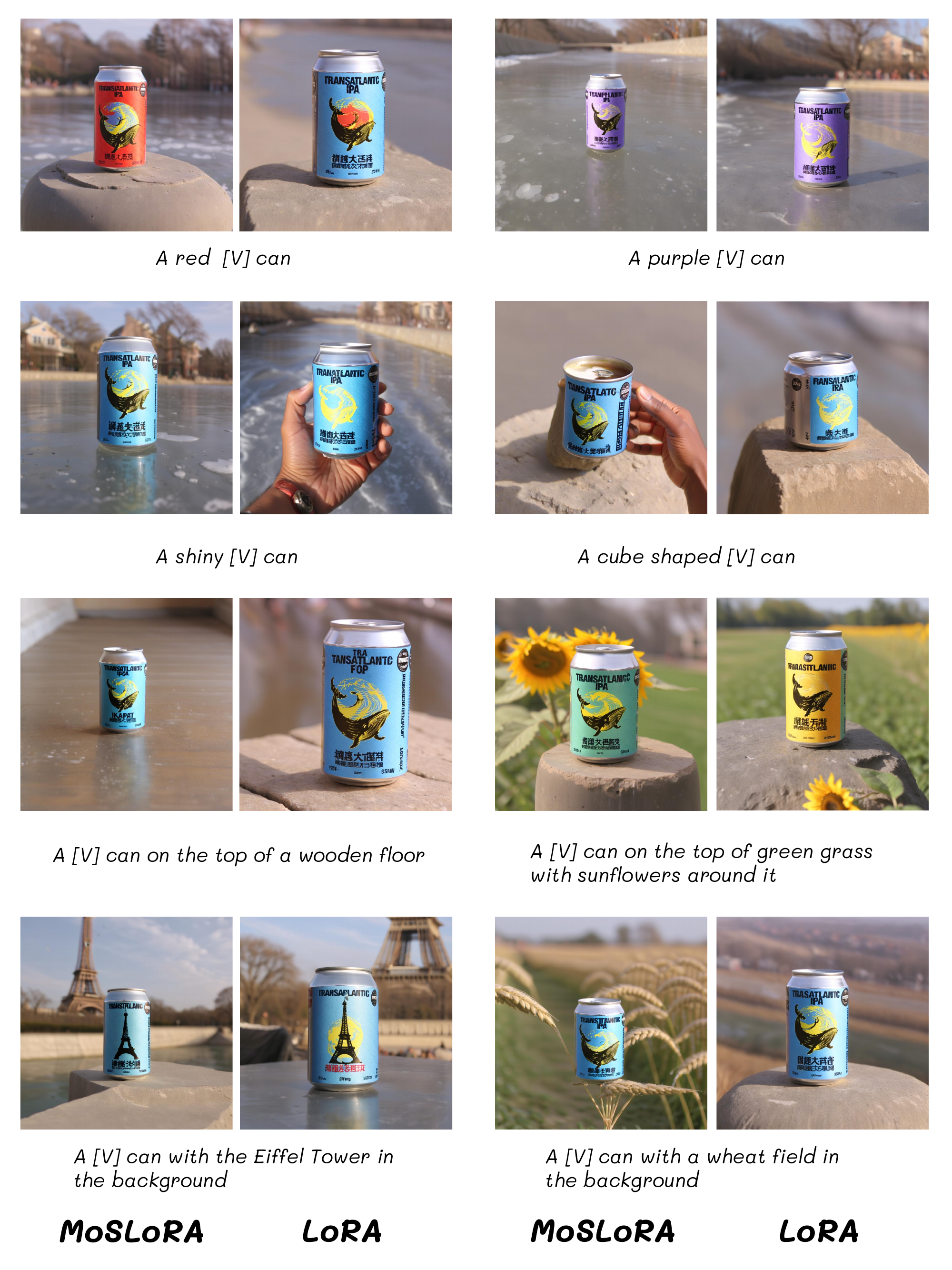}
\caption{
Cases of generated images and paired prompts for the subject \emph{can}. 
}
\label{fig: appendix_sd_cases3}
\end{figure*}

\begin{figure*}[!t]
	\centering
\includegraphics[width=\linewidth]{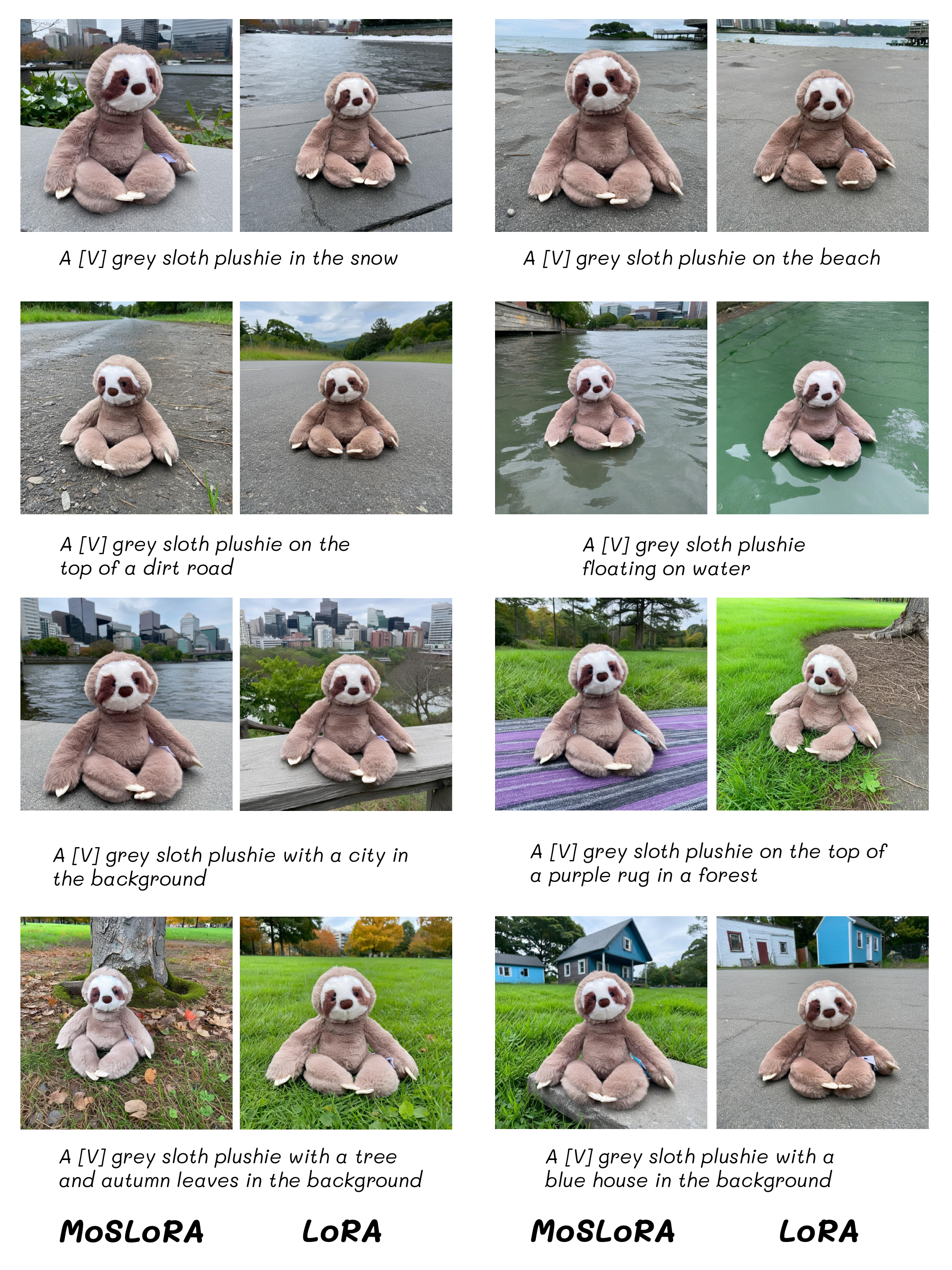}
\caption{
Cases of generated images and paired prompts for the subject \emph{grey sloth plushie}. 
}
\label{fig: appendix_sd_cases4}
\end{figure*}

Figure \ref{fig: appendix_sd_cases1}, \ref{fig: appendix_sd_cases2}, \ref{fig: appendix_sd_cases3}, and \ref{fig: appendix_sd_cases4} show the specific generated images and paired prompts.
For the definition images of these subjects, please refer to the official data\footnote{\href{https://github.com/google/dreambooth/tree/main/dataset}{DreamBooth dataset}} of DreamBooth. 

\onecolumn
\section{Analysis of $\mathbf{W}$ in MoSLoRA}

\subsection{Vanilla MoSLoRA}
For an arbitrary input $x$, we have:
\begin{equation}
\begin{aligned}
    y&=x\mathbf{W}_{merge}; 
    \mathbf{W}_{merge} = \mathbf{W}_0+\mathbf{A}\mathbf{W}\mathbf{B},  
\end{aligned}
\end{equation}
where the $\mathbf{W}_0$ is frozen during training.
Then we have:
\begin{equation}
\begin{aligned}
    \frac{\partial y}{\partial \mathbf{A}}=
    \frac{\partial y}{\partial \mathbf{W}_{merge}} \mathbf{B}^T \mathbf{W}^T;
    \frac{\partial y}{\partial \mathbf{W}}&=\mathbf{A}^T 
    \frac{\partial y}{\partial \mathbf{W}_{merge}} \mathbf{B}^T;
    \frac{\partial y}{\partial \mathbf{B}}=\mathbf{W}^T\mathbf{A}^T 
    \frac{\partial y}{\partial \mathbf{W}_{merge}} 
\end{aligned}
\label{eq: all_gradients}
\end{equation}
Denote the learning rate as $\eta$, the updating process is:
\begin{equation}
\begin{aligned}
 \mathbf{A} \leftarrow \mathbf{A}-\eta \frac{\partial y}{\partial \mathbf{A}} = \mathbf{A}-\eta \frac{\partial y}{\partial \mathbf{W}_{merge}} \mathbf{B}^T \mathbf{W}^T
\end{aligned}
\end{equation}
The process is similar for $\mathbf{W}$ and $\mathbf{B}$.
Let $\Delta= \frac{\partial y}{\partial \mathbf{W}_{merge}}$.
Thus, the weight of the updated LoRA branch would be:
\begin{equation}
\begin{aligned}
    \mathbf{W}_{LoRA} \
&= (\mathbf{A}-\eta \Delta \mathbf{B}^T \mathbf{W}^T)(\mathbf{W}-\eta \mathbf{A}^T 
    \Delta \mathbf{B}^T)
(\mathbf{B}-\eta \mathbf{W}^T\mathbf{A}^T 
    \Delta ) \\
&=(\mathbf{A}\mathbf{W}-\eta \mathbf{A} \mathbf{A}^T 
    \Delta \mathbf{B}^T- \eta \Delta \mathbf{B}^T \mathbf{W}^T \mathbf{W} + \eta^2 \Delta \mathbf{B}^T \mathbf{W}^T \mathbf{A}^T 
    \Delta \mathbf{B}^T
    )
(\mathbf{B}-\eta \mathbf{W}^T\mathbf{A}^T 
    \Delta )
\end{aligned}
\label{lora_old}
\end{equation}

\subsection{Merge $\mathbf{A}$ and $\mathbf{W}$}
Denote $\hat{\mathbf{A}}=\mathbf{A}\mathbf{W}$.
It means that we initialize $\hat{\mathbf{A}}$ as the same as $\mathbf{A}\mathbf{W}$.
The output is the same:
\begin{equation}
\begin{aligned}
    y&=x\mathbf{W}_{merge}; 
    \mathbf{W}_{merge} = \mathbf{W}_0+\hat{\mathbf{A}}\mathbf{B}=\mathbf{W}_0+\mathbf{A}\mathbf{W}\mathbf{B}.
\end{aligned}
\end{equation}
However, the corresponding gradients would be:

\begin{equation}
\begin{aligned}
    \frac{\partial y}{\partial \hat{\mathbf{A}}}= 
    \frac{\partial y}{\partial \mathbf{W}_{merge}} \mathbf{B}^T = \Delta \mathbf{B}^T; 
    \frac{\partial y}{\partial \mathbf{B}}=\hat{\mathbf{A}}^T 
    \frac{\partial y}{\partial \mathbf{W}_{merge}} =  \hat{\mathbf{A}}^T \Delta
\end{aligned}
\end{equation}
Based on that, we can get the updated LoRA after updating the parameters:
\begin{equation}
\begin{aligned}
    \hat{\mathbf{W}}_{LoRA} 
&= (\hat{\mathbf{A}}-\eta \Delta \mathbf{B}^T)
(\mathbf{B}-\eta \hat{\mathbf{A}}^T \Delta ) \\
&= (\mathbf{A}\mathbf{W}-\eta \Delta \mathbf{B}^T)
(\mathbf{B}-\eta \mathbf{W}^T\mathbf{A}^T 
    \Delta ) 
\end{aligned}
\label{lora_new}
\end{equation}

\subsection{Comparison}
Comparing Equation \ref{lora_old} and \ref{lora_new}, we can conclude that the updated weights are not the same, since

\begin{equation}
\begin{aligned}
    \hat{\mathbf{W}}_{LoRA} -  \mathbf{W}_{LoRA}
&= (-\eta \Delta \mathbf{B}^T +\eta \mathbf{A} \mathbf{A}^T 
    \Delta \mathbf{B}^T
+ \eta \Delta \mathbf{B}^T \mathbf{W}^T \mathbf{W} 
- \eta^2 \Delta \mathbf{B}^T \mathbf{W}^T \mathbf{A}^T 
\Delta \mathbf{B}^T)
(\mathbf{B}-\eta \mathbf{W}^T\mathbf{A}^T \Delta ) \\
&= (\eta (\mathbf{A}- \eta \Delta \mathbf{B}^T \mathbf{W}^T ) \mathbf{A}^T \Delta \mathbf{B}^T
+ \eta \Delta \mathbf{B}^T (\mathbf{W}^T \mathbf{W} - \mathbf{I} )
)
(\mathbf{B}-\eta \mathbf{W}^T\mathbf{A}^T \Delta) \neq \mathbf{0}.
\end{aligned}
\end{equation}

\subsection{Fix $\mathbf{W}$ as Orthogonal Matrix}
If we fix $\mathbf{W}$ as \textbf{orthogonal matrix and do not update}~(i.e., $\mathbf{W}\mathbf{W}^{T}=\mathbf{I}$), the updated LoRA would be:
\begin{equation}
\begin{aligned}
    \mathbf{W}^{\mathbf{I}}_{LoRA} \
&= (\mathbf{A}-\eta \Delta \mathbf{B}^T \mathbf{W}^T)\mathbf{W}
(\mathbf{B}-\eta \mathbf{W}^T\mathbf{A}^T 
    \Delta ) \\
&=(\mathbf{A}\mathbf{W}-\eta \Delta \mathbf{B}^T \mathbf{W}^T\mathbf{W})
(\mathbf{B}-\eta \mathbf{W}^T\mathbf{A}^T 
    \Delta) \\
&=(\mathbf{A}\mathbf{W}-\eta \Delta \mathbf{B}^T)
(\mathbf{B}-\eta \mathbf{W}^T\mathbf{A}^T 
    \Delta) = \hat{\mathbf{W}}_{LoRA}
\end{aligned}
\end{equation}

\subsection{Conclusion}

\begin{boxA}
\textit{Though mathematically equivalent initialized, the optimization process would be different if $\mathbf{W}$ is learnable.
Specifically, the optimization process would be the same i.i.f $\mathbf{W}$ is a fixed orthogonal matrix.
}
\end{boxA}

\end{document}